\documentclass[11pt]{article}

\usepackage[final]{acl}
\usepackage{bm}
\usepackage{times}
\usepackage{latexsym}

\usepackage[T1]{fontenc}

\usepackage[utf8]{inputenc}

\usepackage{microtype}
\usepackage{inconsolata}
\usepackage{float} 
\usepackage{hyperref}

\definecolor{lightblue}{RGB}{65, 105, 225}

\hypersetup{
    colorlinks=true,
    linkcolor=lightblue,            
    urlcolor=lightblue        
}

\usepackage{pifont} 

\usepackage{url}
\usepackage{graphicx} 
\usepackage{caption}
\usepackage{booktabs}
\usepackage{multirow}
\usepackage{makecell}
\usepackage{subcaption}
\usepackage{siunitx}
\usepackage{algorithm}
\usepackage{algorithmic}
\usepackage{amsmath, amsthm, amssymb}

\newtheorem{lemma}{Lemma}

\usepackage{wrapfig}
\usepackage{amsfonts}
\theoremstyle{definition}

\usepackage{cleveref}
\usepackage{tabularx}
\theoremstyle{plain}
\usepackage[table]{xcolor}
\usepackage{tikz}
\usetikzlibrary{shapes.misc}
\usepackage{xcolor}     
\definecolor{Gray}{gray}{0.9} 

\usepackage{listings}
\usepackage{pifont}
\usepackage{authblk} 
\usepackage[most,skins,theorems]{tcolorbox}
\tcbset{
  aibox/.style={
    width=\linewidth,
    top=8pt,
    bottom=4pt,
    colback=blue!6!white,
    colframe=black,
    colbacktitle=black,
    enhanced,
    center,
    attach boxed title to top left={yshift=-0.1in,xshift=0.15in},
    boxed title style={boxrule=0pt,colframe=white,},
  }
}
\newtcolorbox{AIbox}[2][]{aibox,title=#2,#1}
\newcommand{\corrauthor}[1]{%
  \begingroup
  \renewcommand{\thefootnote}{\ding{41}}
  \footnotetext{#1}%
  \endgroup
}
\usepackage{pifont}

\title{TokenTiming: A Dynamic Alignment Method for Universal Speculative Decoding Model Pairs}

\author{
  \textbf{Sibo Xiao$^\spadesuit$},
  \textbf{Jinyuan Fu$^\spadesuit$},
  \textbf{Zhongle Xie$^\spadesuit$\textsuperscript{\ding{41}}
  },
  \textbf{Lidan Shou$^{\spadesuit \clubsuit}$},
\\
  $^\spadesuit$Zhejiang University
\\
  $^\clubsuit$Hangzhou High-Tech Zone (Binjiang) Institute of Blockchain and Data Security
\\

    {{\{xiaosibo\_email, 3220100587, xiezl, should\}@zju.edu.cn}}
}

\begin{document}
\maketitle

\begin{abstract}

\corrauthor{Corresponding Author.}

Accelerating the inference of large language models (LLMs) has been a critical challenge in generative AI. Speculative decoding (SD) substantially improves LLM inference efficiency. However, its utility is limited by a fundamental constraint: the draft and target models must share the same vocabulary, thus limiting the herd of available draft models and often necessitating the training of a new model from scratch. Inspired by Dynamic Time Warping (DTW), a classic algorithm for aligning time series, we propose the algorithm TokenTiming for universal speculative decoding. It operates by re-encoding the draft token sequence to get a new target token sequence, and then uses DTW to build a mapping to transfer the probability distributions for speculative sampling. Benefiting from this, our method accommodates mismatched vocabularies and works with any off-the-shelf models without re-training and modification. 
We conduct comprehensive experiments on various tasks, demonstrating 1.57$\times$ speedup. This work enables a universal approach for draft model selection, making SD a more versatile and practical tool for LLM acceleration. The code is available at the \href{https://github.com/SeabirdShore/TokenTiming}{link}.
\end{abstract}

\section{Introduction}
\begin{figure*}[!t]
    \centering
    \includegraphics[width=1\linewidth]{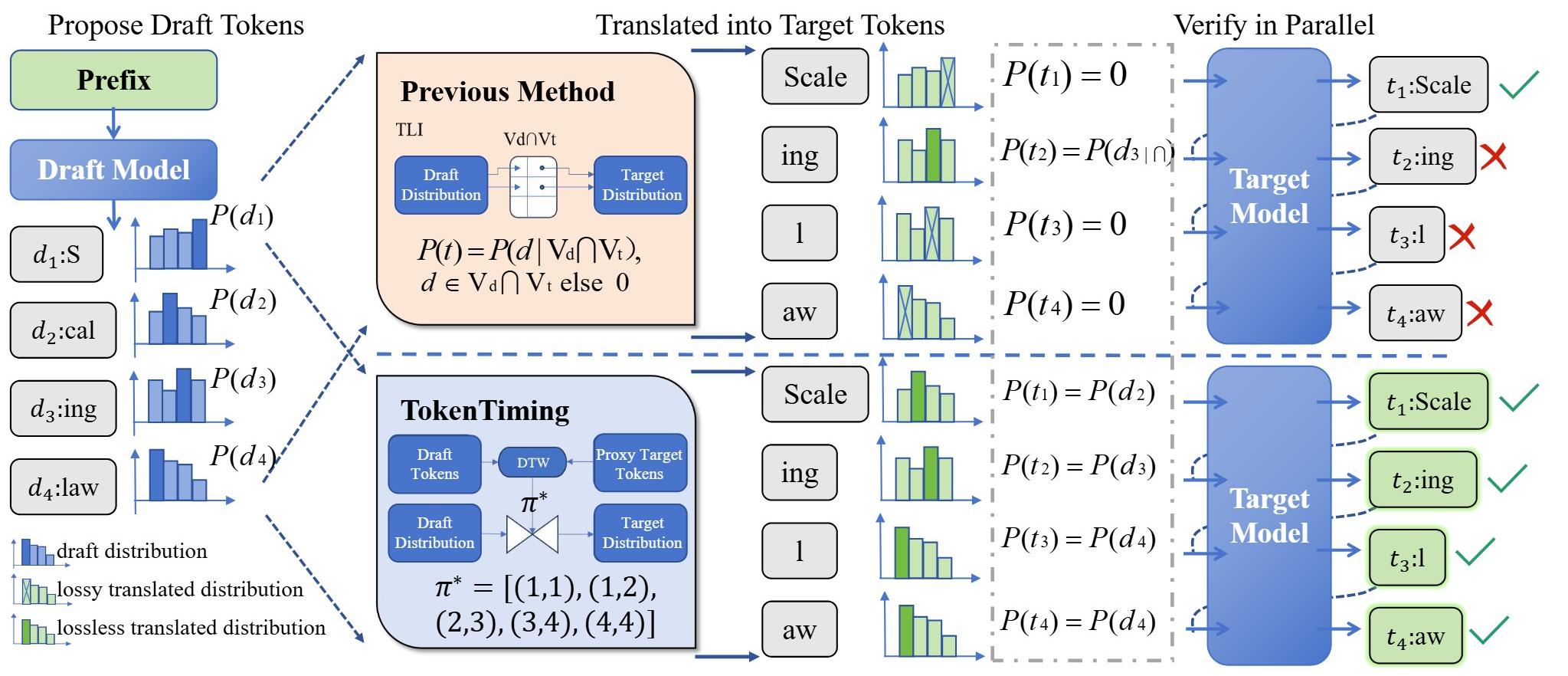}
    \caption{Comparison of core points of TokenTiming with previous work. Previous methods (e.g., TLI) carry the transfer of probability distribution in the vocabulary space of $\mathcal{V}_d \cap \mathcal{V}_t$. If the draft tokens fall outside the intersection, they will be accepted unconditionally, which violates the integrity of losslessness. TokenTiming constructs intact token mapping via DTW, ensuring lossless speculative sampling. }
    \label{fig:overview}
\end{figure*}
Speculative decoding (SD) accelerates LLM inference using a small draft model to propose tokens that are then verified by the larger target model \citep{leviathan2023fast, chen2023accelerating}. 
The effectiveness of SD depends on a draft model that is both fast and accurate in approximating the target distribution \citep{timor2025distributed, chen2024magicdec}. However, a fundamental assumption in current verification methods—the requirement of a shared vocabulary between the draft model and target model—
prevents the widespread adoption of SD \citep{miao2023specinfer, sun2024optimal}. This single constraint leads to two significant practical barriers:
\vspace{-0.7cm}
\paragraph{\textit{Limited Selection of Draft Models.}} SD requires that the target and draft models share the same vocabulary. Many target models are deployed independently, possessing unique vocabularies that fundamentally preclude the choice of draft models. On the other hand, even within the same model family (e.g., GPT-OSS-120B/20B) \citep{openai2025gptoss120bgptoss20bmodel}, the smallest variants often remain too large to provide a substantial draft acceleration. An effective draft model must possess a parameter scale significantly smaller than the target model to ensure high inference efficiency. This dual, stringent constraint on both scale and vocabulary compatibility makes finding the optimal draft model a major practical hurdle. 
\vspace{-0.3cm}
\paragraph{\textit{Costly and Inflexible Training.}}
For a selected target model, obtaining a draft model with complete word alignment usually requires starting from the pre/post-training stage, e.g., Medusa \citep{cai2024medusa}, EAGLE \citep{li2024eagle}. Furthermore, if switched to a new target model, the previously trained model will no longer align. This is extremely inflexible in the current era where there are numerous model types and model iterations occur rapidly.

To address this, several alignment algorithms for universal speculative decoding \citep{timor2025acceleratingllminferencelossless} have been proposed. These methods resolve the heterogeneity of model vocabularies by operating at the linguistic level. SLEM (String-level Exact Match) cannot perform probabilistic sampling, while the performance of TLI (Token-level Intersection) is constrained by the size of the vocabulary intersection between the draft and target models. In conclusion, these methods have only partially alleviated the problem, but they cannot fully meet all the requirements for implementing lossless speculative decoding. 

To overcome these challenges, we present \textbf{TokenTiming}, a novel universal speculative decoding framework that enables lossless acceleration across heterogeneous vocabularies. At the core of TokenTiming is \textbf{Dynamic Token Warping (DTW)}, a lightweight alignment mechanism inspired by Dynamic Time Warping \citep{1163055} from time series analysis. Given a sequence of draft tokens, TokenTiming first converts it into a string and then re-tokenizes it using the target tokenizer to obtain a proxy target token sequence. DTW is then applied to construct a many-to-many alignment between the draft and proxy target token sequence, enabling accurate transfer of probability distributions from the draft vocabulary to the target vocabulary. This alignment is performed \textit{on-the-fly} during each decoding step, without requiring any re-training or model modification. 

\noindent\textbf{Our Contributions:}
\begin{itemize}
\vspace{-0.2cm}
\item \textbf{\textit{Universal Compatibility Without Shared Vocabularies:}} TokenTiming allows any off-the-shelf draft model to be plugged in without a strict vocabulary match.
\vspace{-0.1cm}
\item \textbf{\textit{Strong Empirical Performance Across Tasks:}} On summarization, translation, code, and math, TokenTiming attains up to 1.57× speedup over autoregressive baselines and surpasses universal SD rivals.
\vspace{-0.1cm}
\item \textbf{\textit{Approaching Homogeneous-Vocabulary SD SOTA Performance:}} On 7B/33B Models, TokenTiming yields 2.27× speedup, closing in on Medusa and EAGLE-1/2 while retaining model flexibility.
\end{itemize}
\begin{figure*}[!t]
    \centering
    \includegraphics[width=1\linewidth]{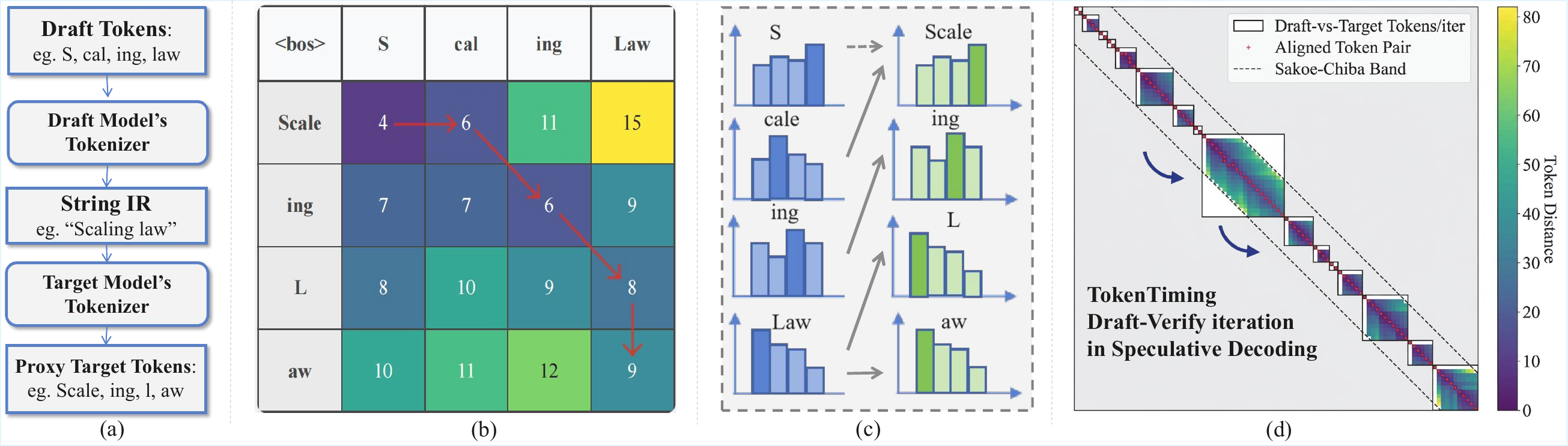}
    \caption{\textbf{Phase illustrations of TokenTiming. (a)} illustrates the re-tokenization of Draft Tokens into Proxy Target Tokens, which are used to construct the mapping in the DTW. \textbf{(b)} DTW calculation process (token distance matrix) and aligned token mapping $\pi ^* =[(S,Scale),(cal,Scale),(ing,ing),(Law,L),(Law,aw)]$. The calculation rules for this mapping are presented in Alg. \ref{alg:dtw_core}. \textbf{(c)} Probability distribution of draft tokens is transferred based on the mapping $\pi^*$. \textbf{(d)} Dynamic alignment in every Draft-Verify iteration of Speculative Decoding.}
    \label{fig:method}
\end{figure*}
\section{Related Works}

\paragraph{SD with Homogeneous Vocabularies}
Speculative Decoding (SD)~\citep{leviathan2023fast, chen2023accelerating} reduces inference latency by using a fast draft model to propose tokens that are then verified in parallel by a larger target model. There are different types of speculative decoding approaches. Draft-head methods like Medusa~\citep{cai2024medusa}, Hydra~\citep{ankner2024hydra}, and EAGLE~\citep{li2024eagle1,li2024eagle,li2025eagle} integrate auxiliary heads into the target model to propose sequences. In contrast, Jacobi-based approaches such as Lookahead Decoding~\citep{fu2024break} and CLLM~\citep{kou2024cllms} enable parallel n-gram generation without draft models. System-level efforts~\citep{miao2023specinfer,liu2024optimizing} further optimize SD's runtime efficiency in serving systems.

\paragraph{SD with Pruned Vocabularies}
\citet{zhao2025fr} proposed a vocabulary pruning strategy to enhance the efficiency of speculative decoding draft models. The core rationale is the pronounced long-tailed structure of token frequency distributions. An empirical analysis of Llama-3-8B on the SlimPajama dataset confirms this, showing that a vast majority of the vocabulary (75\%) accounts for a small fraction (less than 5\%) of token occurrences \citep{grattafiori2024llama, cerebras2023slimpajama}. The proposed method operates by calculating token frequencies on a dataset S and constructing a reduced vocabulary comprising only the most frequent tokens.

\paragraph{SD with Heterogeneous Vocabularies}
Token-Level Intersection (TLI) for heterogeneous draft models has been recently introduced by \citet{timor2025acceleratingllminferencelossless}, which normalizes the intersected distribution by zeroing out all out-of-vocabulary mass. Alternatively, Redistributing draft model Kernels (RDK) \citep{timor2025outofvocabularysamplingboostsspeculative} uses a row-stochastic matrix $M$ to convert the draft distribution $d$ to a new distribution $t'$ via the operation $t' = M^T d$. The resulting distribution $t'$ and the target distribution $t$ are then used for native speculative sampling.

\section{Preliminaries}

\label{sec:prelim}

\paragraph{Speculative Decoding (SD)}
Let $\mathcal V_d$, $\mathcal V_t$ denote the vocabularies of draft model $M_d$ and target model $M_t$ respectively.
SD accelerates autoregressive generation by letting $M_d$ propose $K$ tokens $\mathbf D=(d_1,\dots,d_K)$ in one shot;
$M_t$ then verifies them in a \emph{single} forward pass and rolls back at the first rejection.
The expected speed-up is $\mathbb E[\gamma+1]$ where $\gamma\in[0,K]$ is the number of accepted tokens.
Crucially, standard verification assumes $\mathcal V_d=\mathcal V_t$ so that each draft token $d_i$ can be directly verified by $M_t$.

\paragraph{Standard Speculative Sampling}
Speculative sampling enables \textit{lossless} acceleration by verifying multiple tokens in parallel.  
Let $p(t)$ denote the probability assigned by the draft model to token $t$, and $q(t)$ the corresponding probability under the target model.  
A proposed token $t$ is accepted with probability $\min\!\Bigl(1,\; \frac{q(t)}{p(t)}\Bigr)$, otherwise it is rejected and a new token is sampled from the adjusted target distribution.  
This acceptance rule guarantees that the final output distribution remains identical to the target model's distribution.

\paragraph{Vocabulary Mismatch}
When $\mathcal V_d\neq\mathcal V_t$\footnote{Tab. \ref{tb:result_main_fullwidth} demonstrates detailed statistics of Vocabulary Mismatch by intersection ratio.}, the draft token sequence $\mathbf D$ cannot be directly interpreted by $M_t$:
the same surface form may be tokenised differently (e.g.\ \textbf{``Scaling''} $\to$ one token in $\mathcal V_t$ but \textbf{``Scal''+``ing''} in $\mathcal V_d$),
and normalisation rules make re-encoding non-invertible, so the proxy target token sequence $\mathbf t$ can diverge sharply from $\mathbf D$, collapsing the accept rate.

\section{Method}

To address the fundamental challenges of vocabulary mismatch in universal speculative decoding, we propose \textbf{TokenTiming}, a novel algorithm built upon our core integration of \textbf{Dynamic Token Warping (DTW)}. The complete algorithmic procedure is formally presented in Fig.~\ref{fig:method}.

\subsection{Core Component}

\textbf{Dynamic Token Warping (DTW)} algorithm serves as the core alignment component in our framework, addressing the fundamental challenge of vocabulary mismatch between heterogeneous tokenizers. As formally presented in Alg.~\ref{alg:dtw_core}, DTW establishes an optimal many-to-many mapping between draft token sequence $D=(d_1,\dots,d_k)$ and proxy target token sequence $T=(t_1,\dots,t_m)$ through dynamic programming.

The algorithm operates by constructing a cumulative cost matrix $C \in \mathbb{R}^{k \times m}$, where each entry $C_{i,j}$ represents the minimum cumulative distance to align the first $i$ draft tokens with the first $j$ target tokens. The local dissimilarity $d(x_i, y_j)$ between tokens is computed using an appropriate distance metric, with Levenshtein distance serving as our primary choice due to its effectiveness in capturing token-level edits.

To ensure computational efficiency while maintaining alignment quality, we implement the \textbf{Sakoe-Chiba Band} constraint:
$W = \{(i, j) \mid |i - j| \le w\}$
This restriction limits the search space to a diagonal band of width $w$, reducing computational complexity from $O(k \cdot m)$ to $O(w \cdot \max(k, m))$. The window size $w$ is carefully selected to balance alignment accuracy and computational overhead, with the additional constraint that $\Delta pos < w \land \Delta pos < k$, where $\Delta pos$ denotes the sequence position deviation between matched tokens. The optimal alignment path $\pi^* = [(i_1, j_1), \dots, (i_L, j_L)]$ is obtained through backtracking from $C_{k,m}$ to $C_{1,1}$, following the minimal cost path through the accumulated distance matrix. This path establishes the crucial correspondence that enables subsequent probability transfer between token sequences.

\begin{algorithm}[!t]
\caption{Dynamic Token Warping (DTW)}

\label{alg:dtw_core}
\begin{algorithmic}[1]
    \REQUIRE Token sequences $\mathbf{X}=(x_1, \dots, x_m)$, $\mathbf{Y}=(y_1, \dots, y_n)$, 
             edit distance metric $d$, window size $w$.
    \ENSURE Optimal path $\pi^*$ and total cost $\mathbf{C^*}$.

    \STATE Initialize $C_{0..m, 0..n}$: $C_{0,0} \leftarrow 0$; 
           $C_{i,0}, C_{0,j} \leftarrow \infty$ for $i,j > 0$.

    \FOR{$i \leftarrow 1$ \TO $m$}
        \FOR{$j \leftarrow \max(1, i-w)$ \TO $\min(n, i+w)$}
            \STATE $C_{i,j} \leftarrow d(x_i, y_j) + $
            \STATE $\quad \min\{C_{i-1,j}, C_{i,j-1}, C_{i-1,j-1}\}$
        \ENDFOR
    \ENDFOR

    \STATE $\mathbf{C^*} \leftarrow C_{m,n}$
    \STATE $\pi^* \leftarrow []$, $(i, j) \leftarrow (m, n)$
    \WHILE{$(i, j) \neq (0, 0)$}
        \STATE Prepend $(i, j)$ to $\pi^*$
        \STATE $(i, j) \leftarrow \underset{\substack{(i', j') \in \{ (i-1, j), (i, j-1),  (i-1, j-1) \} }}{\mathrm{argmin} \: C_{i',j'}}$
    \ENDWHILE
    \RETURN $\pi^*$
\end{algorithmic}
\end{algorithm}

\subsection{Pipeline}
\textbf{TokenTiming}, as formally presented in Alg. \ref{alg:speculative_decoding_dtw}, integrates DTW alignment into a cohesive speculative decoding pipeline, enabling efficient cross-tokenizer acceleration while preserving output distribution losslessness.
\subsubsection{Draft Token Calculation}

At each decoding iteration, the draft model $M_d$ autoregressively generates a token sequence $D=(d_1, \dots, d_{k})$ from the current prefix $\mathbf{P}$. To bridge the \textbf{heterogeneous tokenizer} gap, we employ a dual-conversion pipeline: $D$ is first encoded into an intermediate string representation $\mathbf{S}$ using $\text{Tokenizer}_d$, then decoded into the target vocabulary space using $\text{Tokenizer}_t$ to yield \textbf{Proxy Target Tokens} $\mathbf{T}=(t_1, \dots, t_m)$. Notably, the cardinality mismatch $m \ne k$ frequently arises due to fundamental tokenizer disparities, necessitating our DTW alignment mechanism.

\subsubsection{Verification via Speculative Sampling}

Leveraging the mapped probability distribution $\{p(t_j)\}$ derived from DTW alignment, the target model $M_t$ performs parallel verification through a single forward pass over the proposed sequence $T$, computing the true conditional probabilities $\{q(t_j)\}$. Tokens are sequentially accepted according to the speculative decoding criterion:
$\text{Accept } t_j\quad \text{if}\quad r < \min\left(1, \frac{q(t_j)}{p(t_j)}\right)$
where $r \sim U(0, 1)$. This mechanism guarantees that the output distribution exactly matches that of the target model, preserving generative quality while enabling acceleration. The verification process terminates at the first rejection instance, yielding $\gamma$ accepted tokens where $0 \le \gamma \le m$.

\subsubsection{Prefix Update}

Upon accepting $\gamma$ tokens $T^* = (t_{1}, \dots, t_{\gamma})$, the target model generates the subsequent token distribution $P_{M_t}(\cdot \mid \mathbf{P}, t_1, \dots, t_{\gamma})$. We sample $t_{\gamma+1}$ from this distribution and update the decoding prefix as $\mathbf{P} \leftarrow \mathbf{P} \oplus (t_{1}, \dots, t_{\gamma}, t_{\gamma+1})$. This completes one decoding iteration, preparing the system for the next draft generation and verification cycles.

\begin{algorithm}[!t]
    \caption{TokenTiming}
    \label{alg:speculative_decoding_dtw}
    \begin{algorithmic}[1]
        \REQUIRE Prefix $\mathbf{P}$, draft $M_d$, target $M_t$, length $k$
        \ENSURE Full sequence $\mathbf{X}$

        \STATE $\mathbf{X} \leftarrow \mathbf{P}$
        \WHILE{last token of $\mathbf{X} \ne \texttt{<EOS>}$}
            \STATE $D \leftarrow M_{\text{d}}.\text{generate}(\mathbf{X}, k)$
            \STATE $\mathbf{S} \leftarrow \text{Tokenizer}_d.\text{encode}(D)$
            \STATE $T \leftarrow \text{Tokenizer}_t.\text{decode}(\mathbf{S})$
            \STATE $\pi^* \leftarrow \text{DTW}(D,T)$
            
            \STATE $\gamma \leftarrow 0$
            \FOR{$j = 1$ to $|T|$ in parallel}
                \STATE $q(t_j) \leftarrow P_{M_t}(t_j \mid \mathbf{X} \oplus T_{1:j-1})$
                \STATE $p(t_j) \leftarrow \text{MapProbabilities}(p(T), \pi^*, j)$
                \STATE $r \sim U(0, 1)$
                \IF{$r < \min\left(1, q(t_j)/p(t_j)\right)$}
                    \STATE $\gamma \leftarrow \gamma + 1$
                \ELSE
                    \STATE \textbf{break}
                \ENDIF
            \ENDFOR

            \STATE $\mathbf{X} \leftarrow \mathbf{X} \oplus T_{1:\gamma}$
            \STATE $t_{\gamma+1} \leftarrow \text{Sample}(P_{M_t}(\cdot \mid \mathbf{X}))$
            \STATE $\mathbf{X} \leftarrow \mathbf{X} \oplus t_{\gamma+1}$
        \ENDWHILE
        \RETURN $\mathbf{X}$
    \end{algorithmic}
\end{algorithm}

\section{Experiments}
\subsection{Experiment Settings}
\paragraph{LLM Backbones}
To demonstrate our method's effectiveness, we conducted experiments on a diverse set of LLM model pairs, detailed in Tab.~\ref{tab:llm_backbones_compact}. Our selection spans various architectures, sizes, and specializations to ensure a comprehensive evaluation.
The target models include common dense (e.g., Meta-Llama-3.1-70B, Phi-4)\citep{dubey2024llama, abdin2024phi4technicalreport}, distilled (DeepSeek-R1-Distill-Llama-70B, Deepseek-R1-Distill-Qwen-1.5B)\citep{deepseekai2025deepseek}, and Mixture-of-Experts (Qwen3-30B-A3B)\citep{yang2025qwen3technicalreport} architectures, with several being optimized for reasoning. Ranging from 14B to 70B parameters, these models confirm our method's scalability.
For draft models, we showcase the framework's portability by using small, heterogeneous, off-the-shelf models. These include inchoate pre-trained (OPT-350M)\citep{zhang2022opt}, instruction-tuned (Qwen2.5-0.5B), and even an extremely compact fine-tuned model (Vicuna-68M)\citep{zheng2023judging}. The vast size disparity underscores the strong efficiency of our approach, as draft models as small as 68M can effectively accelerate 70B targets.
\paragraph{Generation Settings}
To evaluate our algorithm, we employ Spec-Bench \citep{xia-etal-2024-unlocking}, a comprehensive benchmark designed for assessing Speculative Decoding across diverse scenarios, including translation, summarization, question answering, reasoning, and coding. Spec-Bench integrates CNN/Daily Mail \citep{nallapati-etal-2016-abstractive}, WMT14 DE-EN, Natural Questions \citep{10.1162/tacl_a_00276}, and GSM8K \citep{cobbe2021training} as the primary datasets for these scenarios.
Using this dataset, we perform 480 generations for twenty-five model pairs. All model hyperparameter settings (such as temperature, top\_p, etc.) adopt the default settings from the Hugging Face model library. The detailed hyperparameters for the models used in this experiment are shown in Tab. \ref{tab:cfg}.
\paragraph{Metrics}
 We assess generation efficiency and quality via: (1) \textbf{Tokens Per Second (TPS)}---tokens per second, averaged over the full sequence; (2) \textbf{Accept rate}---fraction of draft tokens accepted by the target model in one speculative step; (3) \textbf{Speedup}---wall-clock time of autoregressive (AR) decoding divided by that of speculative decoding; (4) \textbf{Time to First Token (TTFT)}---time from prompt availability to first token emission; (5) \textbf{Inter-Token Latency (ITL)}---average latency between consecutive tokens; (6) \textbf{Rep-N (Repetition-N)}---proportion of duplicate $n$-grams among all $n$-grams in the output, which measures diversity \citep{shao2019longdiversetextgeneration}.

\subsection{Experiment Results}

\subsubsection{Overall Results}
Tab.~\ref{tb:result_main_fullwidth} presents a comprehensive evaluation of our proposed method, TokenTiming, against autoregressive (AR) decoding and a recent speculative decoding algorithm, TLI \citep{timor2025acceleratingllminferencelossless}\footnote{RDK \citep{timor2025distributed} is not included in the baseline due to the inaccessible code source.}, which is designed for heterogeneous vocabularies. The experiments span a diverse set of large-scale target models, including DeepSeek-R1-Distill-Llama-70B model, Llama-3.1-70B, Qwen3-30B, Qwen3-32B, and Phi-4, paired with various small draft models such as Deepseek-R1-Distill-Qwen-1.5B (abbreviated as DQwen-1.5B in Tab. \ref{tb:result_main_fullwidth}). Performance is primarily measured by the speedup ratio relative to the AR baseline and the absolute TPS.

\begin{table*}[t!]
\centering
{\footnotesize
\renewcommand{\arraystretch}{0.8}
\setlength{\tabcolsep}{3pt}
\setlength{\aboverulesep}{0.5pt}
\setlength{\belowrulesep}{1pt}
\setlength{\cmidrulesep}{0.5pt}
\setlength{\extrarowheight}{-1pt}

\definecolor{ar-gray}{RGB}{240,240,240}
\definecolor{blue-light}{RGB}{225,240,255}
\definecolor{blue-medium}{RGB}{180,220,255}

\sisetup{
  round-mode=places,
  round-precision=2,
}

\begin{tabular}{l l c c S[table-format=2.2] S[table-format=2.2] S[table-format=1.2] S[table-format=1.2] S[table-format=1.2] S[table-format=1.2]}
\toprule
\toprule

\multirow{2}{*}{\textbf{Target Model}} & \multirow{2}{*}{\textbf{Draft Model}} & \multirow{2}{*}{$\bm{\frac{\mathcal V_d \cap \mathcal V_t }{\mathcal V_d \cup \mathcal V_t}}$} & \multirow{2}{*}{$\bm{\frac{\mathcal V_d \cap \mathcal V_t }{\max\{\mathcal V_d, \mathcal V_t\}}}$} & \multicolumn{2}{c}{\textbf{TPS}} & \multicolumn{2}{c}{\textbf{Accept Rate}} & \multicolumn{2}{c}{\textbf{Speedup}} \\
\cmidrule(l){5-6} \cmidrule(l){7-8} \cmidrule(l){9-10}
& & & & {\textbf{TLI}} & {\textbf{Ours}} & {\textbf{TLI}} & {\textbf{Ours}} & {\textbf{TLI}} & {\textbf{Ours}} \\
\midrule

\multirow{5}{*}{\makecell[l]{DeepSeek-R1\\-Distill-Llama-70B}}
 & Autoregressive & \cellcolor{ar-gray}{-} & \cellcolor{ar-gray}{-} & \multicolumn{2}{c}{\cellcolor{ar-gray}14.68} & \cellcolor{ar-gray}{-} & \cellcolor{ar-gray}{-} & \cellcolor{ar-gray}{-} & \cellcolor{ar-gray}{-} \\
 & \makecell[l]{Qwen2.5-0.5B} & 0.643 & 0.722 & 15.68 &  \textbf{19.26}   & 0.34 &  \textbf{0.40}   & 1.068120 &  \textbf{1.31}   \\
 & Qwen3-0.6B & 0.643 & 0.722 & 14.14 &   15.35 & 0.28 &   0.29 & 0.963215 &   1.045640 \\
 & DQwen-1.5B &0.643&0.722&17.76&  19.08&0.29&  0.37&1.21& 1.30\\
 & Vicuna-68M & 0.064 & 0.075 & 14.79 &   15.43 & 0.23 &   0.26 & 1.007493 &   1.051090 \\
 & OPT-350M & 0.319 & 0.337 & 16.03 &  \textbf{21.35}   & 0.19 &  \textbf{0.31}   & 1.091962 &  \textbf{1.45}   \\
\cmidrule{1-10}

\multirow{5}{*}{\makecell[l]{Llama-3.1-70B}}
 & Autoregressive & \cellcolor{ar-gray}{-} & \cellcolor{ar-gray}{-} & \multicolumn{2}{c}{\cellcolor{ar-gray}13.55} & \cellcolor{ar-gray}{-} & \cellcolor{ar-gray}{-} & \cellcolor{ar-gray}{-} & \cellcolor{ar-gray}{-} \\
 & \makecell[l]{Qwen2.5-0.5B} & 0.643 & 0.722 & 16.35 &   14.38 & 0.34 &   0.31 & 1.206642 &   1.061255 \\
 & Qwen3-0.6B & 0.643 & 0.722 & 14.18 &   15.03 & 0.23 &  0.33 & 1.046494 &   1.109225 \\
 & DQwen-1.5B &0.643&0.722&16.67& \textbf{18.25} &0.20&  0.22&1.23& 1.35 \\
 & Vicuna-68M & 0.064 & 0.075 & 15.58 &  16.53 & 0.19 &   0.06 & 1.149815 &  1.219926 \\
 & OPT-350M & 0.319 & 0.337 & 16.56 &  \textbf{17.84}   & 0.35 &   \textbf{0.25}  & 1.222140 &  \textbf{1.32}   \\
\cmidrule{1-10}

\multirow{5}{*}{\makecell[l]{Qwen3-30B-A3B}}
 & Autoregressive & \cellcolor{ar-gray}{-} & \cellcolor{ar-gray}{-} & \multicolumn{2}{c}{\cellcolor{ar-gray}9.80} & \cellcolor{ar-gray}{-} & \cellcolor{ar-gray}{-} & \cellcolor{ar-gray}{-} & \cellcolor{ar-gray}{-} \\
 & \makecell[l]{Qwen2.5-0.5B} & 0.999 & 0.999 & 10.74 &   11.37 & 0.36 &  0.37 & 1.095918 &   1.160204 \\
 & Qwen3-0.6B & 1.000 & 1.000 & 11.71 &  \textbf{11.90}   & 0.44 &  \textbf{0.45}   & 1.194898 &  \textbf{1.21}   \\
 & DQwen-1.5B & 0.999 & 0.999 & 12.23 &  \textbf{12.90}   & 0.41 &  \textbf{0.45}   & 1.25 &  \textbf{1.32}   \\
 & Vicuna-68M &0.055&0.063&8.33&  10.89&0.26&  0.34&0.85&  1.11\\
 & OPT-350M & 0.265 & 0.279 & 9.99 &   9.95 & 0.23 &   0.33 & 1.019388 &   1.015306 \\
\cmidrule{1-10}

\multirow{5}{*}{Qwen3-32B}
 & Autoregressive & \cellcolor{ar-gray}{-} & \cellcolor{ar-gray}{-} & \multicolumn{2}{c}{\cellcolor{ar-gray}15.77} & \cellcolor{ar-gray}{-} & \cellcolor{ar-gray}{-} & \cellcolor{ar-gray}{-} & \cellcolor{ar-gray}{-} \\
 & \makecell[l]{Qwen2.5-0.5B} & 0.999 & 0.999 & 20.97 &  \textbf{24.47}   & 0.42 &  0.38 & 1.329740 &  \textbf{1.55}   \\
 & Qwen3-0.6B & 1.000 & 1.000 & 19.16 &  \textbf{24.78}   & 0.43 &  0.42 & 1.214965 &  \textbf{1.57}   \\
 & DQwen-1.5B & 0.999 & 0.999 & 18.83 &  \textbf{21.90}   & 0.35 &  0.39   & 1.194 &  \textbf{1.39}   \\
 & Vicuna-68M &0.055&0.063&14.67&  18.92&0.19& 0.33 &0.93&  1.20\\
 & OPT-350M & 0.265 & 0.279 & 17.28 &  \textbf{24.01}   & 0.18 &   \textbf{0.21}  & 1.095751 &  \textbf{1.52}   \\
\cmidrule{1-10}

\multirow{5}{*}{Phi-4}
 & Autoregressive & \cellcolor{ar-gray}{-} & \cellcolor{ar-gray}{-} & \multicolumn{2}{c}{\cellcolor{ar-gray}23.14} & \cellcolor{ar-gray}{-} & \cellcolor{ar-gray}{-} & \cellcolor{ar-gray}{-} & \cellcolor{ar-gray}{-} \\
 & \makecell[l]{Qwen2.5-0.5B} & 0.649 & 0.654 & 17.64 &  \textbf{35.58}   & 0.19 &  \textbf{0.39}   & 0.762316 &  \textbf{1.54}   \\
 & Qwen3-0.6B & 0.649 & 0.654 & 26.80 &   28.59 & 0.34 &  0.27 & 1.158168 &  1.235523 \\
 & DQwen-1.5B &0.649&0.654&22.91&  24.07&0.15&  0.19&0.99&  1.04\\
 & Vicuna-68M & 0.078 & 0.095 & 22.19 &   28.09 & 0.13 &   0.05 & 0.958946 &  1.213915 \\
 & OPT-350M & 0.400 & 0.429 & 31.64 &   28.33 & 0.21 &   0.20 & 1.367329 &  1.224287 \\

\bottomrule
\bottomrule
\end{tabular}
}
\caption{
  Performance comparison between TokenTiming and baselines, AR and TLI \citep{timor2025acceleratingllminferencelossless}, across a diverse set of target and draft models.
  The primary metrics are the speedup ratio relative to AR and absolute Tokens Per Second (TPS).
  The results demonstrate TokenTiming's significant and consistent performance superiority over TLI.
  Comparatively optimal results for each target model are highlighted in \textbf{bold}.}
\label{tb:result_main_fullwidth}

\end{table*}

\begin{table*}[t!]
\centering

{\footnotesize
\renewcommand{\arraystretch}{0.8}
\setlength{\tabcolsep}{3pt}
\setlength{\aboverulesep}{0.5pt}
\setlength{\belowrulesep}{1pt}
\setlength{\cmidrulesep}{0.5pt}
\setlength{\extrarowheight}{-1pt}

\begin{tabular}{l l *{5}{S[table-format=1.2, round-precision=2, detect-weight] S[table-format=1.2, round-precision=2, detect-weight]}}
\toprule
\toprule

\multirow{2}{*}{\textbf{Target Model}} & \multirow{2}{*}{\textbf{Draft Model}} & \multicolumn{2}{c}{\textbf{Math}} & \multicolumn{2}{c}{\textbf{Program}} & \multicolumn{2}{c}{\textbf{Translation}} & \multicolumn{2}{c}{\textbf{Summarize}} & \multicolumn{2}{c}{\textbf{QA}} \\
\cmidrule(lr){3-4} \cmidrule(lr){5-6} \cmidrule(lr){7-8} \cmidrule(lr){9-10} \cmidrule(l){11-12}
& & {\textbf{TLI}} & {\textbf{Ours}} & {\textbf{TLI}} & {\textbf{Ours}} & {\textbf{TLI}} & {\textbf{Ours}} & {\textbf{TLI}} & {\textbf{Ours}} & {\textbf{TLI}} & {\textbf{Ours}} \\
\midrule

\multirow{4}{*}{\makecell[l]{DeepSeek-R1\\-Distill-Llama-70B}}
 & \makecell[l]{Qwen2.5-0.5B} & 1.03 & \bfseries 1.48 & 1.05 & \bfseries 1.34 & 1.62 & \bfseries 2.54 & 0.94 & \bfseries 1.07 & 0.99 & \bfseries 1.08 \\
 & Qwen3-0.6B                 & 0.95 & \bfseries 1.34 & 0.88 & \bfseries 1.35 & 1.01 & \bfseries 2.53 & 1.02 & \bfseries 1.06 & 1.11 & 1.08 \\
 & DQwen-1.5B                 & 1.05 & \bfseries 1.58 & 0.99 & \bfseries 1.24 & 1.10 & \bfseries 1.65 & 1.05 & \bfseries 1.31 & 1.03 & \bfseries 1.53 \\
 & Vicuna-68M                 & 0.97 & \bfseries 0.98 & 0.81 & \bfseries 1.04 & 1.22 & \bfseries 1.44 & 1.02 & \bfseries 1.30 & 1.05 & \bfseries 1.22 \\
 & OPT-350M                   & 1.02 & \bfseries 1.18 & 1.05 & \bfseries 1.11 & 0.96 & \bfseries 2.05 & 1.01 & \bfseries 1.09 & 0.97 & \bfseries 1.08 \\
\cmidrule{1-12}

\multirow{4}{*}{\makecell[l]{Llama-3.1-70B}}
 & \makecell[l]{Qwen2.5-0.5B} & 0.92 & \bfseries 1.34 & 0.88 & \bfseries 1.47 & 1.09 & \bfseries 1.10 & 0.89 & \bfseries 1.06 & 1.51 & 1.32 \\
 & Qwen3-0.6B                 & 1.04 & \bfseries 1.57 & 1.30 & 1.17 & 0.73 & \bfseries 1.54 & 0.86 & \bfseries 0.91 & 1.0 & \bfseries 1.11 \\
 & DQwen-1.5B                 & 0.95 & \bfseries 1.43 & 1.21 & \bfseries 1.24 & 1.15 & \bfseries 1.64 & 1.11 & \bfseries 1.34 & 1.22 & \bfseries 1.64 \\
 & Vicuna-68M                 & 1.21 & \bfseries 1.25 & 1.05 & \bfseries 1.36 & 0.84 & \bfseries 1.22 & 1.07 & \bfseries 1.22 & 1.04 & \bfseries 1.21 \\
 & OPT-350M                   & 0.94 & \bfseries 1.36 & 1.14 & \bfseries 1.44 & 0.99 & \bfseries 1.33 & 1.12 & \bfseries 1.14 & 1.26 & \bfseries 1.32 \\
\cmidrule{1-12}

\multirow{4}{*}{\makecell[l]{Qwen3-30B-A3B}}
 & \makecell[l]{Qwen2.5-0.5B} & 1.15 & \bfseries 1.62 & 0.76 & \bfseries 1.23 & 1.31 & \bfseries 1.45 & 1.21 & \bfseries 2.03 & 0.71 & \bfseries 1.04 \\
 & Qwen3-0.6B                 & 1.24 & 1.05 & 0.92 & \bfseries 1.25 & 1.21 & 1.06 & 0.82 & \bfseries 1.11 & 0.57 & \bfseries 0.85 \\
 & \makecell[l]{DQwen-1.5B}   & 1.14 & \bfseries 1.35 & 0.92 & \bfseries 1.32 & 1.11 & \bfseries 1.16 & 0.91 & \bfseries 1.10 & 0.88 & \bfseries 0.95 \\
 & Vicuna-68M                 & 1.02 & \bfseries 1.35 & 1.13 & \bfseries 1.36 & 0.95 & \bfseries 1.52 & 1.25 & \bfseries 1.44 & 1.03 & \bfseries 1.34 \\
 & OPT-350M                   & 1.06 & 1.04 & 0.84 & \bfseries 1.02 & 0.75 & \bfseries 0.91 & 0.96 & \bfseries 1.18 & 0.74 & \bfseries 0.77 \\
\cmidrule{1-12}

\multirow{4}{*}{Qwen3-32B}
 & \makecell[l]{Qwen2.5-0.5B} & 1.44 & \bfseries 2.53 & 1.29 & \bfseries 1.80 & 1.91 & \bfseries 2.49 & 1.41 & 1.28 & 1.01 & \bfseries 1.13 \\
 & Qwen3-0.6B                 & 1.16 & \bfseries 2.05 & 1.01 & \bfseries 2.47 & 1.50 & \bfseries 1.61 & 1.35 & \bfseries 1.80 & 1.44 & 1.41 \\
 & \makecell[l]{DQwen-1.5B}   & 1.13 & \bfseries 1.15 & 1.12 & \bfseries 1.15 & 1.01 & \bfseries 1.12 & 0.97 & \bfseries 1.14 & 0.97 & \bfseries 1.05 \\
 & Vicuna-68M                 & 0.83 & \bfseries 1.45 & 0.93 & \bfseries 1.39 & 0.94 & \bfseries 1.44 & 1.25 & \bfseries 1.63 & 1.06 & \bfseries 1.44 \\
 & OPT-350M                   & 1.16 & \bfseries 1.92 & 1.77 & \bfseries 2.16 & 1.66 & 1.24 & 0.91 & \bfseries 1.05 & 0.91 & \bfseries 1.52 \\
\cmidrule{1-12}

\multirow{4}{*}{Phi-4}
 & \makecell[l]{Qwen2.5-0.5B} & 0.90 & \bfseries 1.57 & 0.94 & \bfseries 1.60 & 0.67 & \bfseries 1.31 & 0.71 & \bfseries 1.55 & 0.76 & \bfseries 1.55 \\
 & Qwen3-0.6B                 & 1.04 & \bfseries 1.51 & 1.21 & \bfseries 1.61 & 1.05 & \bfseries 1.20 & 1.09 & 1.07 & 1.48 & 1.21 \\
 & \makecell[l]{DQwen-1.5B}   & 1.21 & \bfseries 1.45 & 1.32 & \bfseries 1.75 & 1.06 & \bfseries 1.52 & 1.07 & \bfseries 1.74 & 1.12 & \bfseries 1.45 \\
 & Vicuna-68M                 & 0.87 & \bfseries 1.52 & 0.83 & \bfseries 0.88 & 0.74 & \bfseries 1.41 & 0.87 & \bfseries 1.64 & 1.29 & \bfseries 1.30 \\
 & OPT-350M                   & 1.64 & 1.34 & 1.30 & \bfseries 1.41 & 1.43 & \bfseries 1.52 & 1.59 & 1.42 & 1.43 & \bfseries 1.55 \\

\bottomrule
\bottomrule
\end{tabular}
}
\caption{Speedup comparison between TokenTiming and baseline across different task categories. The table shows results for Math, Programming, Translation, Summarization, and Question Answering tasks. Results where our method outperforms the baseline are highlighted in \textbf{bold}.}
\label{tb:result_task_categories}
\end{table*}

\begin{figure*}[!t]
\begin{minipage}[h]{0.34\textwidth} 
    \centering
    \begin{figure}[H] 
        \includegraphics[width=\linewidth]{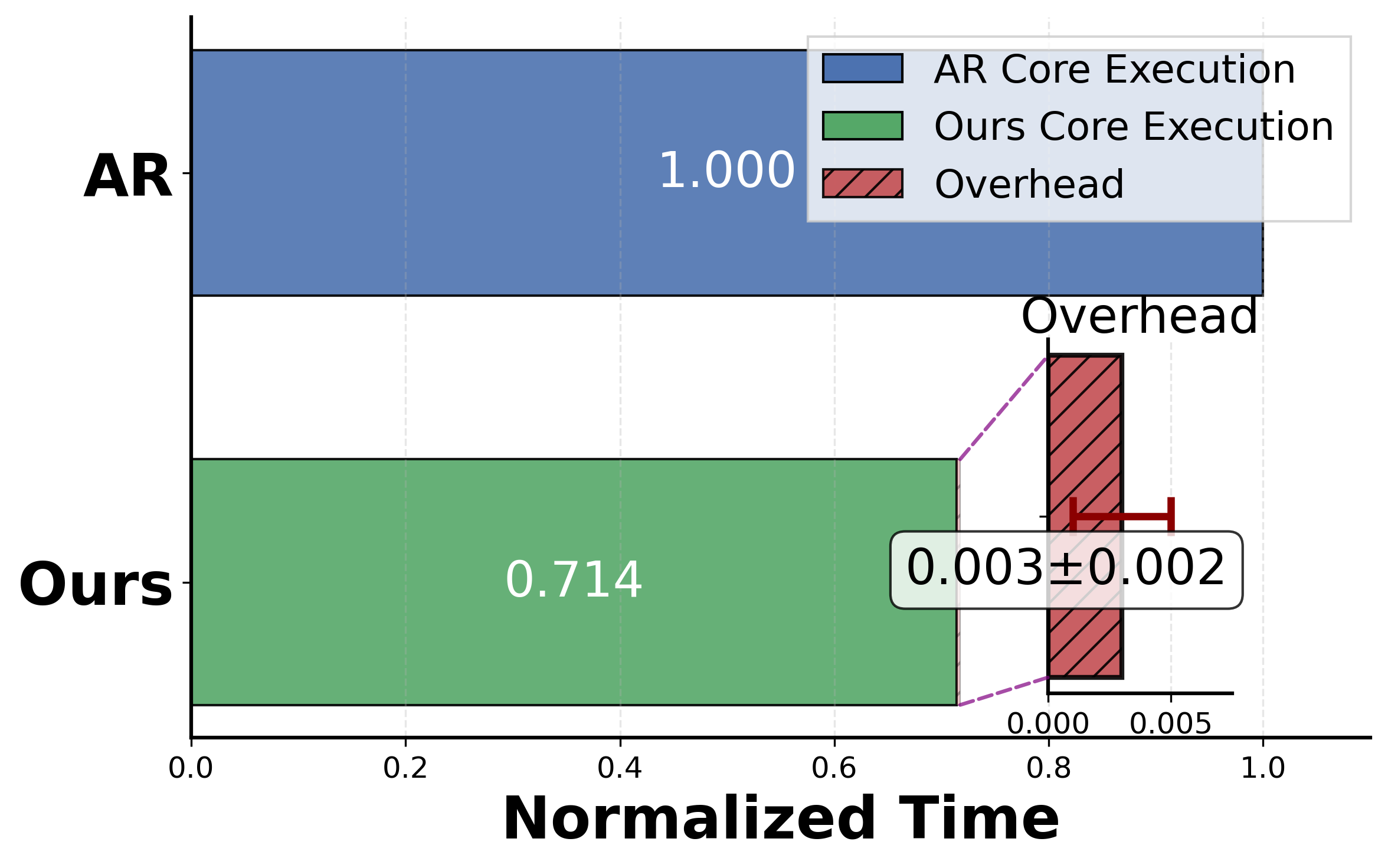}
        \caption{Total Time Cost (with TokenTiming overhead)}
        \label{fig:overhead}
    \end{figure}
\end{minipage}%
\hfill
\begin{minipage}[h]{0.61\textwidth} 
    \centering
    \begin{figure}[H] 
        \includegraphics[width=\linewidth]{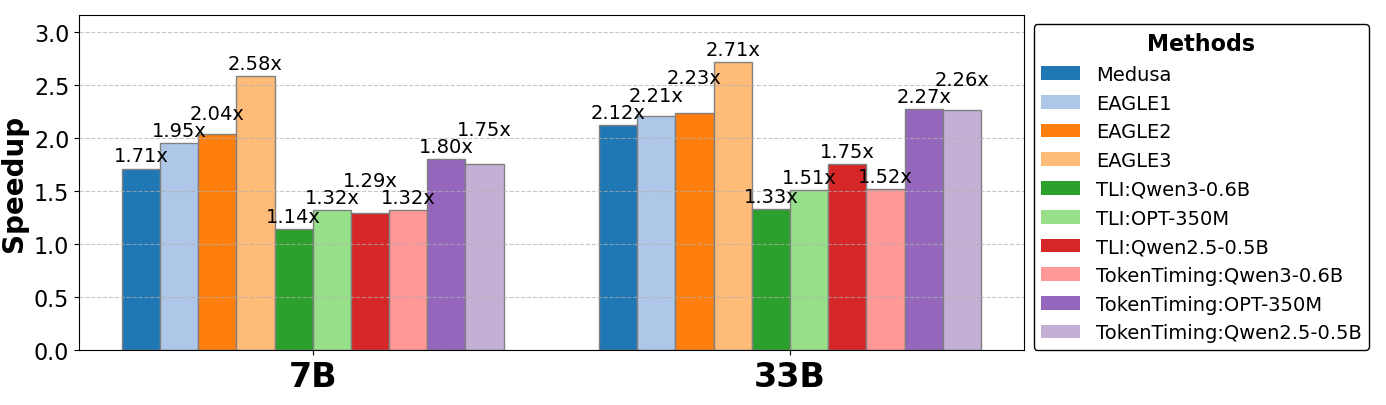}
        \caption{Speed-up vs.\ homogeneous-vocabulary SOTA on various target model scales and draft models. TokenTiming bridges the performance gap while keeping the universal draft-model advantage.}
        \label{fig:sota}
    \end{figure}
\end{minipage}
\end{figure*}
The results demonstrate the superior performance of our proposed TokenTiming algorithm. Across all tested target models, TokenTiming consistently achieves a higher speedup than both the AR baseline and the TLI method. For instance, when accelerating the Qwen3-32B model, TokenTiming achieves a remarkable speedup of up to 1.57$\times$ (using the Qwen3-0.6B draft model), which is a significant improvement over the maximum speedup of 1.33$\times$ achieved by TLI. Similarly, for the DeepSeek-R1-Distill-Llama-70B model, TokenTiming reaches a speedup of 1.45$\times$, whereas TLI's peak performance is capped at 1.09$\times$.
Furthermore, the analysis highlights the critical role of draft model selection in speculative decoding. Our results indicate that TokenTiming is highly effective at leveraging different draft models to maximize performance. For example, with the Llama-3.1-70B target, TokenTiming paired with OPT-350M yields a 1.32$\times$ speedup. For the Phi-4 model, TokenTiming achieves its peak performance of 1.54$\times$ speedup with the Qwen2.5-0.5B draft model, again substantially outperforming TLI's best result of 1.37$\times$.

This enhanced speedup is often correlated with a robust token accept rate. In many configurations, TokenTiming not only delivers higher TPS but also maintains a competitive or even higher accept rate than TLI. This suggests that TokenTiming's mechanism is more efficient at generating candidate sequences that are accepted by the target model, leading to more effective acceleration.
\subsubsection{Approaching Homogeneous-Vocabulary SD Acceleration}

\paragraph{\textit{Comparable Performance to Homogeneous-Vocabulary SOTA.}}
As shown in Fig. \ref{fig:sota}, on 7B-target model, the best homogeneous-vocabulary SD method (EAGLE-3) reaches 2.58× speed-up, while the strongest heterogeneous-vocabulary baseline (TLI) peaks at 1.32×.
TokenTiming with OPT-350M closes this gap to 1.80×, only 0.78× away from EAGLE-3.
On 33B-target model the trend is identical: TokenTiming-OPT-350M delivers 2.27×, which is within 0.44× of EAGLE-3 (2.71×) and already surpasses Medusa (1.71×) and EAGLE-1 (2.21×).

\paragraph{\textit{More Draft Model Choices without Re-training.}}  
Medusa/EAGLE requires extra parameters integrated into the target model or costly draft-head re-training when the target is switched.  
TokenTiming is plug-and-play: any off-the-shelf 68M–350M model can be used while instantaneously yielding 1.05×–2.27× speed-up across four diverse draft models, demonstrating the flexibility that homogeneous-vocabulary SOTA cannot provide.

\subsubsection{Optimization on DTW Constraints}
\begin{figure}[!t]
    \centering 
    \includegraphics[width=\linewidth]{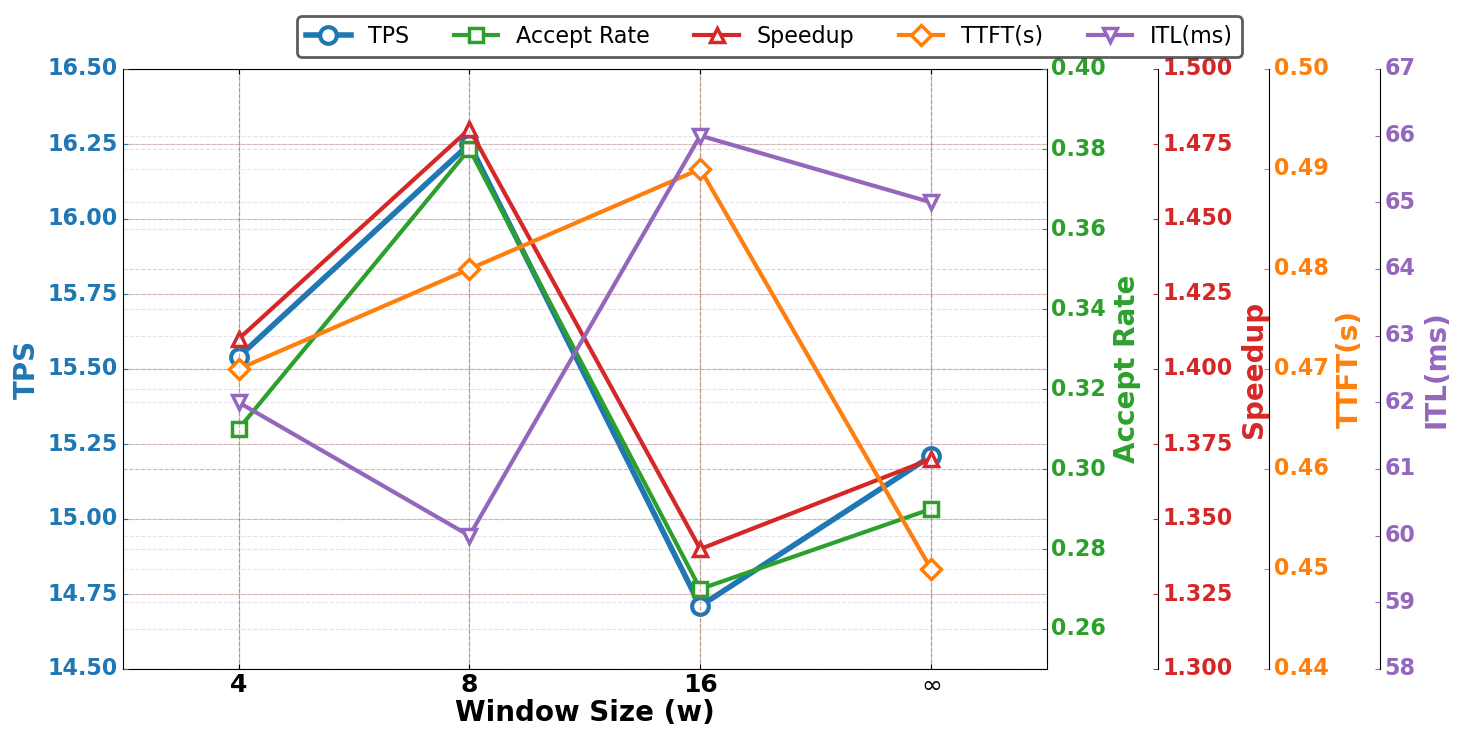} 
    \caption{Performance metrics under various settings of $w$: $w=4$, $w=8$, $w=16$, and $w=\infty$.} 
    \label{fig:radar} 
\end{figure}

\begin{figure}[!t]
    \centering 
    \includegraphics[width=0.90\linewidth]{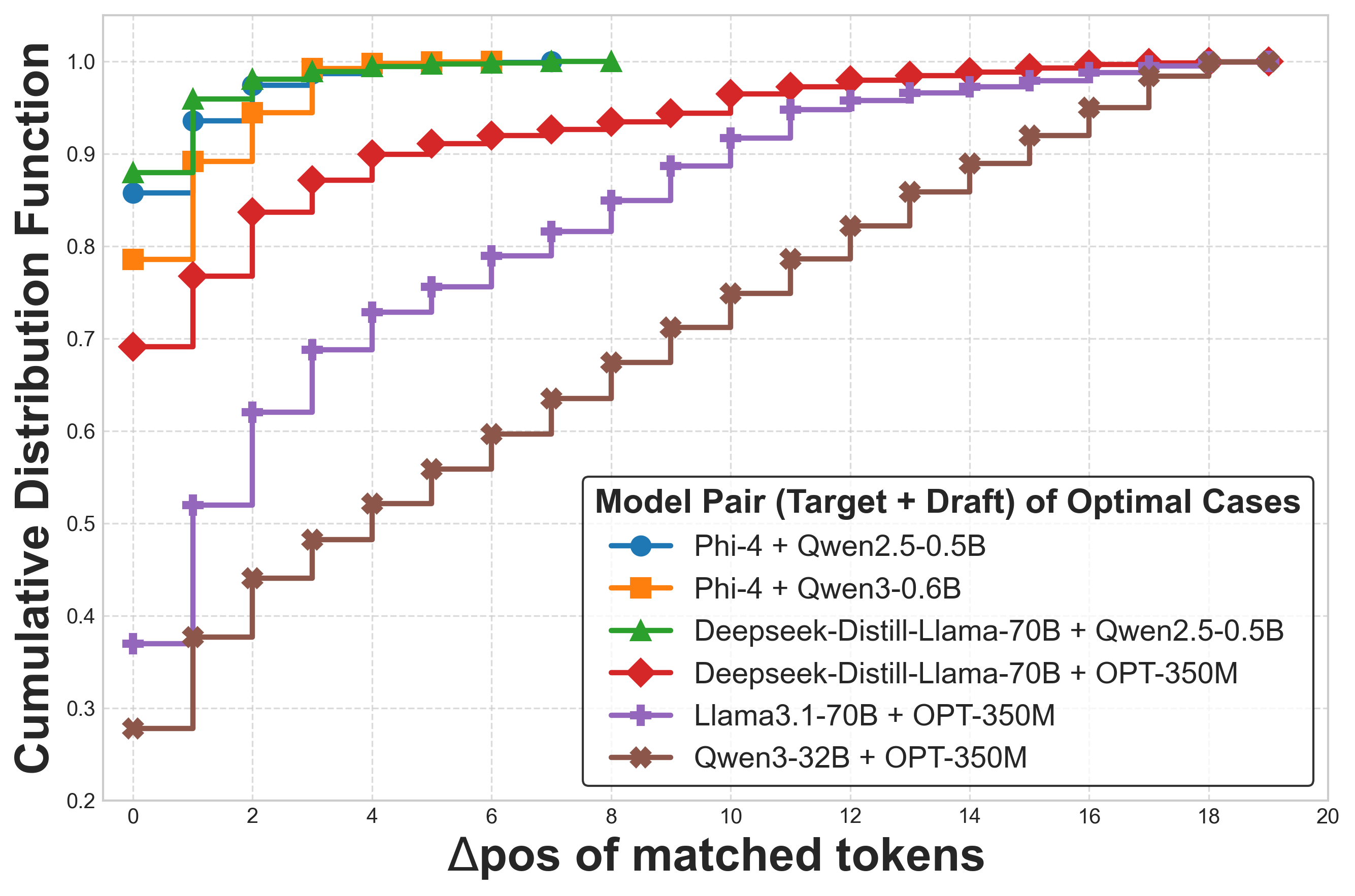} 
    \caption{Cumulative Distribution Function of $\Delta pos$  of DTW matched tokens. $w$ = $\infty$. $\Delta pos$  of DTW matched tokens converges to an upper bound.} 
    \label{fig:cdf} 
\end{figure}

We conducted experiments under different settings of $w$: $w=4$, $w=8$, $w=16$, and $w=\infty$, where $w=\infty$ corresponds to the case without the Sakoe-Chiba Band. The results are shown in the Fig.~\ref{fig:radar}. By selecting an appropriate window size $w$, the acceleration effect of TokenTiming on SD can be partially improved, resulting in higher TPS, Accept Rate, and Speedup.
As shown in Fig.~\ref{fig:cdf}, the upper bound of $\Delta pos$ convergence exceeds the setting of our best-performing $w=8$ in some model pairs, and the proportion of cases where $\Delta pos > 8$ is non-trivial. However, compared to $w=\infty$, $w=8$ still achieves performance improvement. This result indicates that appropriate constraints can better preserve the probability distribution of tokens.

\subsubsection{Task-Specified Performance}
Tab.~\ref{tb:result_task_categories} compares the task-level speedup of TokenTiming against TLI.TokenTiming consistently outperforms TLI on all five evaluated tasks.
For mathematics, TokenTiming reaches 2.53× with the pair Qwen3-32B + Qwen2.5-0.5B, while TLI only achieves 1.44×.
Similar gaps appear in summarization (2.54× vs. 1.62×) and translation (1.60× vs. 0.94×).
Although the absolute speedup on programming and QA is slightly lower, TokenTiming still maintains a noticeable margin.

The gain is tightly correlated with the capability of the draft model.
Strong draft models such as Qwen2.5-0.5B and Qwen3-0.6B boost the accept rate and push speedup beyond 2× on reasoning-intensive tasks.
Conversely, light-weight drafts (OPT-350M, Vicuna-68M) reduce the advantage, especially on math and code generation where token-level accuracy is crucial.
Nevertheless, even with these weaker drafts, TokenTiming remains superior to TLI, confirming the robustness of the DTW alignment when vocabulary overlap is limited.

\subsubsection{Tradeoff Analysis}

Fig. \ref{fig:overhead} shows that the integration of Dynamic Token Warping into the decoding pipeline introduces an additional computational step\footnote{In Appendix \ref{appendix:trace}, it is shown that for the CPU/GPU stream timeline, the blocking time introduced by TokenTiming in one decoding cycle is only 663$\mu s$, which is trivial compared to other parts.}, creating a negligible time overhead  (0.1\% to 0.5\% of overall runtime) compared to simpler, direct token-matching methods. This overhead is an inherent trade-off for the enhanced flexibility and accuracy of the alignment process. However, our empirical analysis indicates that this additional cost is effectively amortized by the substantial gains in overall generation throughput (1.4$\times$ speedup). 


\subsubsection{DTW Bridges the Gap of Mismatched Vocabularies}
Fig.~\ref{fig:cdf} shows that the sequence position deviation of matched proxy target token and draft token is not exclusively zero, where position means the positional order in the two token sequence. If the vocabularies of the two models were perfectly aligned, the DTW matching result would converge precisely on the diagonal. The presence of this deviation, therefore, demonstrates that our algorithm is effectively matching these imperfectly aligned vocabularies. We observe that a significant proportion of the deviation is non-zero, which highlights a key advantage of the DTW algorithm over the TLI algorithm. TokenTiming is capable of handling skewed probability distribution mappings, rather than one-to-one mappings, in a sequential manner, thereby increasing the accept rate.
\subsubsection{Eliminate Potential Inflated Performance}
\begin{figure}[!t]
    \centering
    \includegraphics[width=1\linewidth]{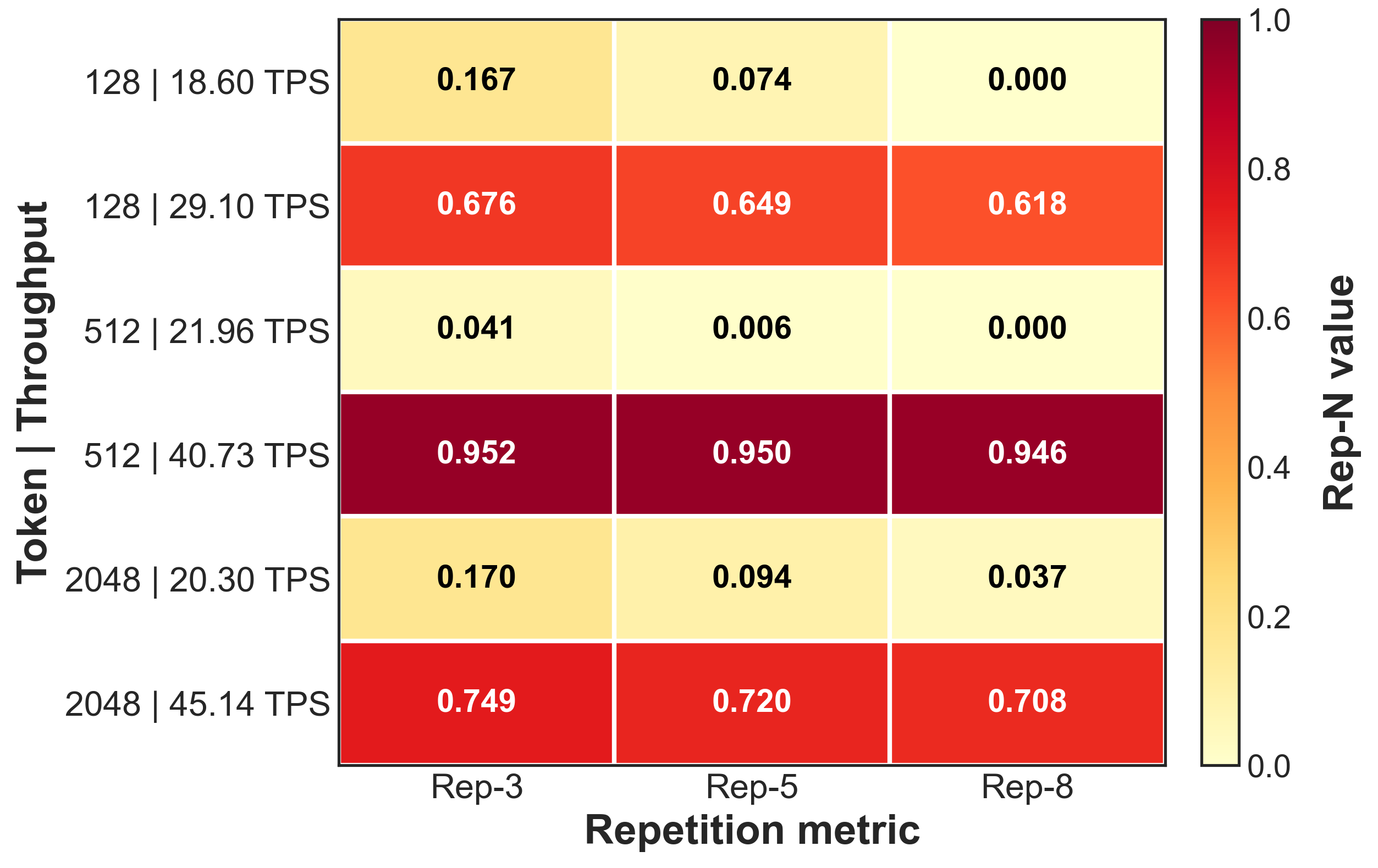}
    \caption{Repetition-N \citep{shao2019longdiversetextgeneration} in settings with a different number of generated tokens.}
    \label{fig:rep}
\end{figure}

While our results demonstrate significant speedup, scenarios such as repetitive generation loops can artificially inflate this metric. In these cases, the draft model easily predicts subsequent repetitive tokens, leading to a near-perfect accept rate and a high number of accepted tokens per step. This corresponds to technically fast but non-meaningful generation, as shown in Fig.~\ref{fig:rep}. To avoid distortion, we exclude such pathological repetitions (approximately \textbf{15\%} of test samples) from our main analysis, following prior work showing they can amplify token probabilities and inflate speed metrics~\citep{xu2022learningbreakloopanalyzing}. The remaining 85\% of samples, covering standard tasks, thus provide a representative assessment of typical generation behavior. 

\subsubsection{Modes on Highly-fragmented Tokenizations}
\begin{figure}[!t]
    \centering
    \includegraphics[width=1\linewidth]{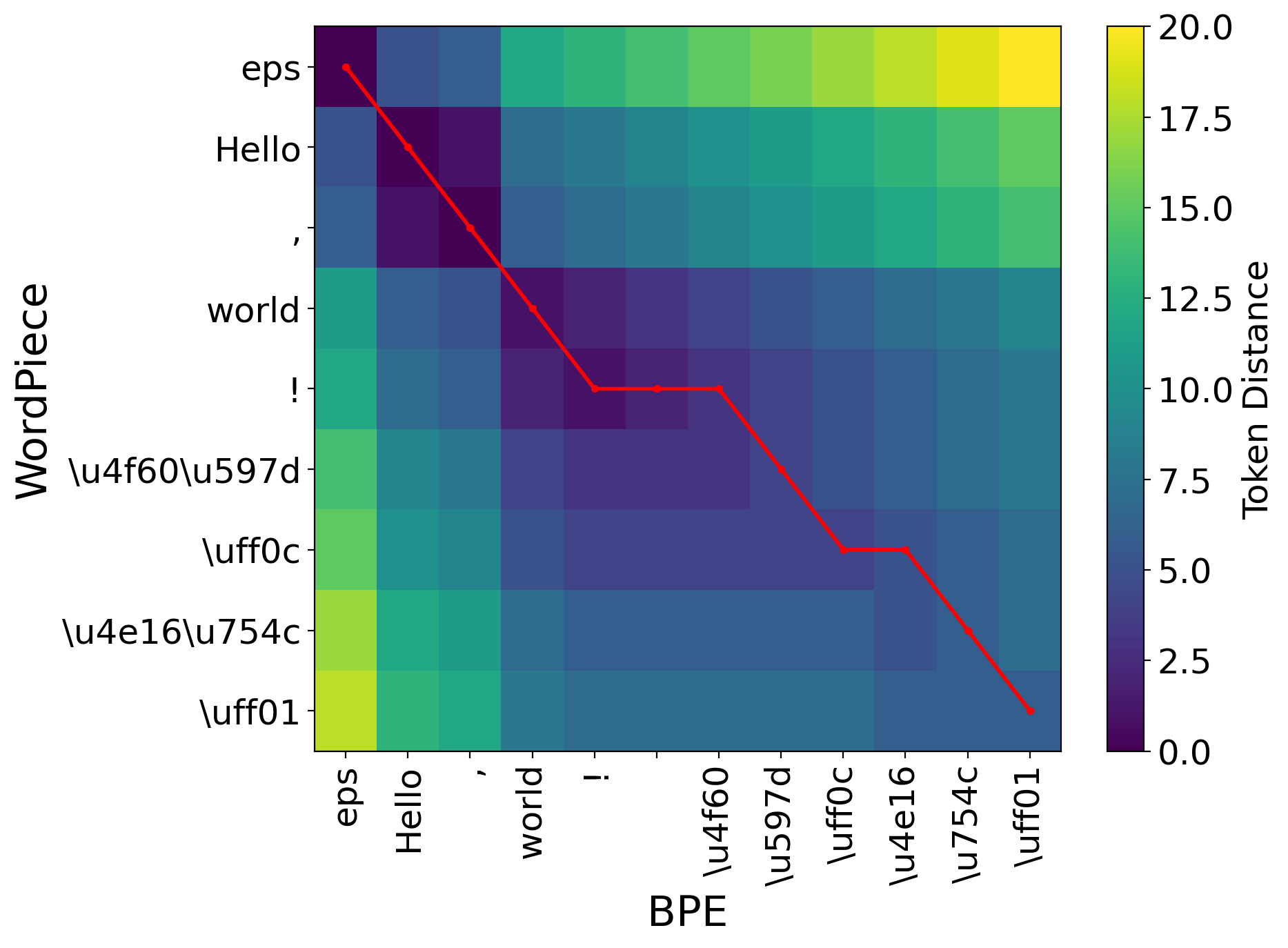}
    \caption{Alignment path for fragmentarily tokenized prompt. The prompt was constructed by inserting special symbols which have different tokenized results between the draft and target tokenizers and which are highly-fragmented at least in one tokenizer.}
    \label{fig:pt}
\end{figure}

Fig.~\ref{fig:pt} shows the alignment paths of the tokenized results for the prompt with special symbols by two common tokenization methods, WordPiece and Byte-Pair Encoding(BPE) \cite{sennrich2016neural}. We can observe that the special symbols are converted into the UTF8 encoding format when used for calculating Token Distance. The UTF8 format has a minimum resolution of one byte during tokenization \cite{wang2019neuralmachinetranslationbytelevel}, and it has strong specificity in the alignment path. Even if the granularity of tokenization for special characters by the two tokenizers is different, the token at the end of the two tokenization sequences is still aligned. This is consistent with the probability transfer assignment logic of TokenTiming, that is, the probability of the end token of the draft token sequence block and the target token sequence block is equal. The alignment result demonstrates robustness.
\vspace{-0.2cm}

\section{Conclusion}

This paper introduced TokenTiming, a dynamic alignment algorithm that eliminates the shared-vocabulary constraint of existing speculative decoding methods. By employing a dynamic warping approach to build a lossless probability mapping between heterogeneous vocabularies, TokenTiming enables any off-the-shelf model to serve as a draft model without re-training. Experiments show our method achieves up to a 1.57$\times$ speedup over previous state-of-the-art speculative decoding algorithms across various generation tasks. By removing the fundamental bottleneck of heterogeneous vocabularies, TokenTiming makes speculative decoding a more versatile and practical tool for LLM acceleration.

\section*{Limitations}
The probability form calculated in TokenTiming is one-hot, i.e., it directly transfers the probability of the top-1 token. The calculation of DTW is based on the character granularity rather than the semantic granularity. There are certain differences in the alignment effect between languages with different word tokenization granularities.

\section*{Ethics Statement}
TokenTiming focuses on expanding the application scope of speculative decoding models. The related research involves time series alignment and language tokenization techniques, aiming to enhance the decoding efficiency of large language models. There are no technical security risks or risks of technical abuse.

\section*{Acknowledgements}
We sincerely thank the reviewers, the area chairs, and the program chairs for their constructive comments. This work was supported by the Pioneer R\&D Program of Zhejiang (No. 2024C01021) and Zhejiang Province ``Leading Talent of Technological Innovation Program'' (No. 2023R5214).

\bibliography{custom}
\appendix
\section{Consistency and Losslessness Proof}
Before proceeding to the core proof, we must formalize the process by which the draft probability distribution $p(t)$ is generated and prove its consistency across mismatched vocabularies. Let $\mathcal V_d$ and $\mathcal V_t$ be the vocabularies for the draft and target models, respectively. The draft model, $M_d$, generates a sequence of tokens $D = (d_1, \dots, d_K)$ where $d_i \in V_d$. This output is transformed into a distribution $p(t)$ over $\mathcal V_t$ through a two-stage procedure:

\subsection{Deterministic Re-tokenization and String Consistency}
The first stage involves a mapping $F_{\text{retokenize}}: \mathcal V_d^* \to \mathcal V_t^*$. Let $f_{\text{decode}}^d$ and $f_{\text{encode}}^t$ be the decoding function of the draft and the encoding function of the target, respectively.
\begin{lemma}[String-level Consistency]
The proxy target token sequence $T = (t_1, \dots, t_m) = f_{\text{encode}}^t(f_{\text{decode}}^d(D))$ is identical to the draft sequence $D$ in the string space.
\end{lemma}
\begin{proof}
Let $S = f_{\text{decode}}^d(D)$ be the string generated by the draft model. Modern tokenizers (e.g., BPE, SentencePiece) satisfy the \textbf{lossless invertibility} property for any valid UTF-8 string: $f_{\text{decode}}^t(f_{\text{encode}}^t(S)) \equiv S$. 
Substituting the definition of $T$:
\[
f_{\text{decode}}^t(T) = f_{\text{decode}}^t(f_{\text{encode}}^t(S)) = S = f_{\text{decode}}^d(D)
\]
Thus, while the token boundaries differ ($K \neq m$), the underlying character streams are identical, ensuring the verification process targets the exact same content.
\end{proof}

\subsection{Losslessness of Probability Mapping}
The alignment $\pi^* = \text{DTW}(D, T)$ maps indices of $D$ to $T$. We define the proposal probability $p(t_j)$ based on the following two non-trivial mapping cases:

\paragraph{Case 1: Many-to-One (Draft  ``a'',  ``b''$\to$ Target  ``ab'').} 
In this case, multiple draft tokens $\{d_i, \dots, d_{i+n}\}$ correspond to a single target token $t_j$. Our implementation assigns the probability of the \textbf{terminal} draft token to the target token: $p(t_j) = P_d(d_{i+n} \mid \text{prefix}, d_i, \dots, d_{i+n-1})$.
\begin{proof}
In the target model, $q(t_j)$ represents the probability of completing the string unit ``ab'' given the prefix. In the draft model, since the prefix and the initial part of the unit ``a'' are already assumed/fixed during the sequential alignment, the probability of the final token $d_{i+n}$ ``b'' represents the draft's confidence in \textit{completing} that specific string block. Under the assumption that $M_d$ approximates $M_t$, the conditional probability of completing a semantic unit is a consistent estimator of the target's atomic probability for that unit.
\end{proof}

\paragraph{Case 2: One-to-Many (Draft ``ab'' $\to$ Target ``a'', ``b'').} 
Here, one draft token $d_i$ covers multiple target tokens $\{t_j, \dots, t_{j+k}\}$. Our implementation \textbf{copies} the probability: $p(t_j) = \dots = p(t_{j+k}) = P_d(d_i)$.
\begin{proof}
Since $d_i$ is a single token in $\mathcal V_d$, $P_d(d_i)$ represents the draft's total confidence in the string ``ab''. 
1) For the first segment $t_j$ (``a''), $P_d(d_i)$ acts as a conservative surrogate. Since ``a'' is a prefix of ``ab'', $P(\text{``a''}) \geq P(\text{``ab''})$ holds in a joint distribution. Using $p(t_j) = P_d(d_i)$ ensures $p$ is not overestimated relative to the target's marginal.
2) For subsequent segments like $t_{j+1}$ (``b'' given ``a''), the draft model has already committed to the full string ``ab''. Therefore, the probability of the component ``b'' appearing is implicitly 1.0 within that draft step. By using the original $P_d(d_i)$, we maintain a high proposal probability that matches the target model's conditional confidence, maximizing the accept rate $\min(1, q/p)$.
\end{proof}

\section{Example of Generated Text}
\label{appendix-b}

\begin{tcolorbox}[
    enhanced,
    breakable,
    colback=blue!5!white,
    colframe=blue!75!black,
    fonttitle=\bfseries,
    watermark color=blue!10!white
]
Draft model: DeepSeek-R1-Distill-Qwen-1.5B,
Target model: DeepSeek-R1-Distill-Llama-70B,
Prefix: The field of Artificial Intelligence has seen tremendous growth in recent years. One of the most exciting areas of research is
\tcblower 
The field of Artificial Intelligence has seen tremendous growth in recent years. One of the most exciting areas of research is Reinforcement Learning (RL), which involves training agents to make decisions in complex, uncertain environments. In this blog post, we will explore the concept of policy in reinforcement learning, its types, and the importance of policy optimization.

\#\#\# What is a Policy in Reinforcement Learning?

In reinforcement learning, a **policy** is a strategy that an agent uses to decide actions based on the current state of the environment. It can be thought of as a function, often denoted as $\pi(a|s)$, which takes the current state (s) and returns an action (a) or a probability distribution over possible actions. The goal of the agent is to learn an optimal policy that maximizes the cumulative reward over time.

\#\#\# Types of Policies

Policies in RL can be categorized in different ways. Here are some common types:

1. **Deterministic Policy**: 
   - A deterministic policy always selects the same action for a given state. It is a direct mapping from states to actions. For example,$\pi$(s) = a.

2. **Stochastic Policy**:
   - A stochastic policy, on the other hand, selects actions probabilistically. It outputs a probability distribution over possible actions given a state. This is often useful in exploration-exploitation trade-offs, where the agent might sometimes choose a suboptimal action to discover better ones.

3. **Parametric Policy**:
   - Parametric policies are defined by a set of parameters. These parameters can be adjusted during training to improve the policy. Examples include neural networks, where the weights and biases are the parameters.

4. **Non-Parametric Policy**:
   - Non-parametric policies do not rely on a fixed set of parameters. Instead, they might be represented by lookup tables or other structures that can grow with the data. These are less common in deep RL settings.
   
\end{tcolorbox}

\section{Hardware Configuration Specification}
The hardware configuration of our experiment environments can be seen in Tab.~\ref{tab:hardware_specs}. 
\begin{table}[!h]
\centering
\begin{tabular}{ll}
\toprule
\textbf{Component} & \textbf{Specification} \\
\midrule
\textbf{CPU} & 2 x Intel Xeon Platinum 8558 \\
 & \quad (Total 96 Cores / 192 Threads) \\
\addlinespace
\textbf{RAM} & 2.0 TiB \\
\addlinespace
\textbf{GPU} & 2 x NVIDIA H100 80GB HBM3 \\
 & - Architecture: Hopper \\
 & - VRAM per GPU: 80 GB \\
 & - Total VRAM: 160 GB\\
 & - Interconnect: PCIe Gen 5 x16 \\
\addlinespace
\textbf{Software} 
 &  NVIDIA Driver: 570.133.20 \\
 &  CUDA Version: 12.8 \\
\bottomrule
\end{tabular}
\caption{Hardware Environment  Configuration}
\label{tab:hardware_specs}
\end{table}
\section{Related Algorithms}
\subsection{Token-level Intersection}
Token-Level Intersection, Alg.~\ref{alg:vocabs-intersection}, for heterogeneous draft models has been recently introduced by \citet{timor2025acceleratingllminferencelossless}, which normalizes the intersected distribution by zeroing out all out-of-vocabulary mass in the probability conversion.
\begin{algorithm}[!t]
    \caption{(Token-Level Intersection, TLI \citep{timor2025acceleratingllminferencelossless}), an iteration of speculative decoding for heterogeneous vocabularies with token-level rejection sampling verification}
    \label{alg:vocabs-intersection}
    \begin{algorithmic}[1]
    \STATE \textbf{Input:} Probability distributions $p$ and $q$ over vocabularies $T$ and $D$, respectively. Drafting lookahead $i \in \mathbb{N}$.

    \STATE \textbf{Output:} A sequence of tokens from $T$, containing between $1$ and $i+1$ tokens.
    \STATE \textbf{Procedure:}
    \STATE Define a probability distribution $q'$ over the vocabulary $T \cap D$ such that $q'(x)=\frac{q(x)}{\sum_{t \in T} q(t)}$ if $x \in T$ and $q'(x) = 0$ otherwise.
    \STATE \textbf{Run} Standard Speculative Decoding with $p, q', i, c$.
    \end{algorithmic}
\end{algorithm}
\subsection{Standard Speculative Decoding}
Standard Speculative Decoding \cite{leviathan2023fast, chen2023accelerating}, Alg.~\ref{alg:sd}, accelerates autoregressive generation by pre-generating candidate tokens with a draft model for batch verification, converting sequential decoding into parallel computation. The algorithm ensures distribution consistency via probability ratio acceptance but is limited to homogeneous vocabularies, which TokenTiming overcomes through dynamic alignment.

\begin{algorithm}[!t]
    \caption{Standard Speculative Decoding \citep{leviathan2023fast,chen2023accelerating}}
    \label{alg:sd}
    \begin{algorithmic}[1]
    \STATE \textbf{Input:} Probability distributions $p$ and $q$ over a vocabulary $T$. Drafting lookahead $i \in \mathbb{N}$. An input prompt~$c$.
    \STATE \textbf{Output:} A sequence of tokens from $T$, containing between $1$ and $i+1$ tokens.
    \STATE \textbf{Procedure:}
    \FOR{$j \leftarrow 1, \ldots, i$}
        \STATE Sample a draft token from the drafter, $d_j \sim q_{c \oplus d_1 \oplus \ldots \oplus d_{j-1}}$.
    \ENDFOR
    \STATE In parallel, compute the $i+1$ logits of the target model: $p_{c},~p_{c \oplus d_1},~\cdots,~p_{c \oplus d_1 \oplus \cdots \oplus d_i}$.
    \FOR{$j \leftarrow 1, \ldots, i$}
        \STATE Let $x \leftarrow c \oplus d_1 \oplus \cdots \oplus d_{j-1}$.
        \IF{$p_{x}(d_j) \le q_{x}(d_j)$}
            \STATE With probability $1 - \frac{p_x(d_j)}{q_x(d_j)}$, \textbf{reject} $d_j$ and \textbf{break}. 
        \ENDIF
        \STATE \textbf{Accept} the draft token $d_j$.
    \ENDFOR
    \STATE Let $j \in \{0, 1, \ldots, i\}$ be the number of accepted drafts. \label{line:count-accepted-drafts}
    \STATE Set $x \leftarrow c \oplus d_1 \oplus \ldots \oplus d_j$.
    \STATE Sample $t \sim r_{x}$ where $r_x(\cdot)$ is the modified distribution.
    \STATE \textbf{Return} $d_1, \ldots, d_j, t$.
    \end{algorithmic}
\end{algorithm}
\section{LLM Backbones}
The experiment selects LLMs with diverse architectures and parameter scales as target and draft models, enabling universal validation of TokenTiming through diversified model configurations. The detailed model configurations are listed in Tab.~\ref{tab:llm_backbones_compact}.
\begin{table}[!t]
\centering
\scriptsize 
\setlength{\tabcolsep}{4pt} 
\begin{tabular}{llc}
\toprule
\textbf{Model} & \textbf{Type} & \textbf{Params} \\ 
\midrule
\multicolumn{3}{c}{\textit{Target Models}} \\
\midrule
Llama-3.1-70B & Dense & 70B \\
DeepSeek-R1-Distill-LLama-70B & Distill, Reasoning & 70B \\
Phi-4 & Dense & 14B \\
Qwen3-32B & Dense, Reasoning & 32B \\
Qwen3-30B-A3B & MoE, Reasoning & 30B \\
\midrule
\multicolumn{3}{c}{\textit{Draft Models}} \\
\midrule
Vicuna-68M & Fine-Tuned & 68M \\
Qwen3-0.6B & Dense, Reasoning & 0.6B \\
Qwen2.5-0.5B(-Instruct) & Instruction-Tuned & 0.5B \\
OPT-350M & Dense & 350M \\
\bottomrule
\end{tabular}
\caption{Overview of LLM backbones used in our experiments. The selection covers a wide range of model types and sizes. Roles are indicated by subheadings.}
\label{tab:llm_backbones_compact}
\end{table}
\begin{figure*}[!t]
    \centering
    \includegraphics[width=1\linewidth]{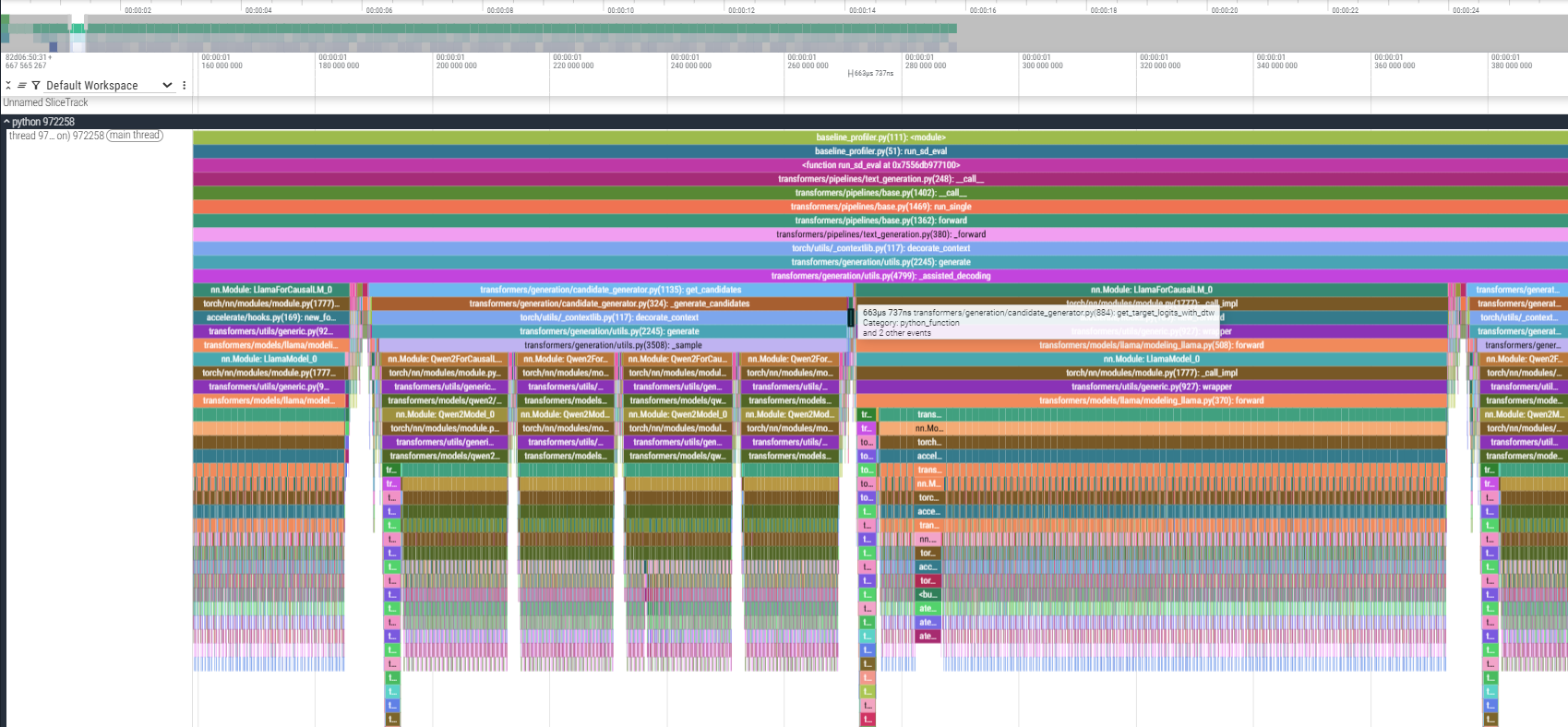}
    \caption{Trace visualization. The process segment for target logits calculation with DTW-related operations takes only 663$\mu s$, highlighting its extremely short duration.}
    \label{fig:cpu trace}
    \includegraphics[width=1\linewidth]{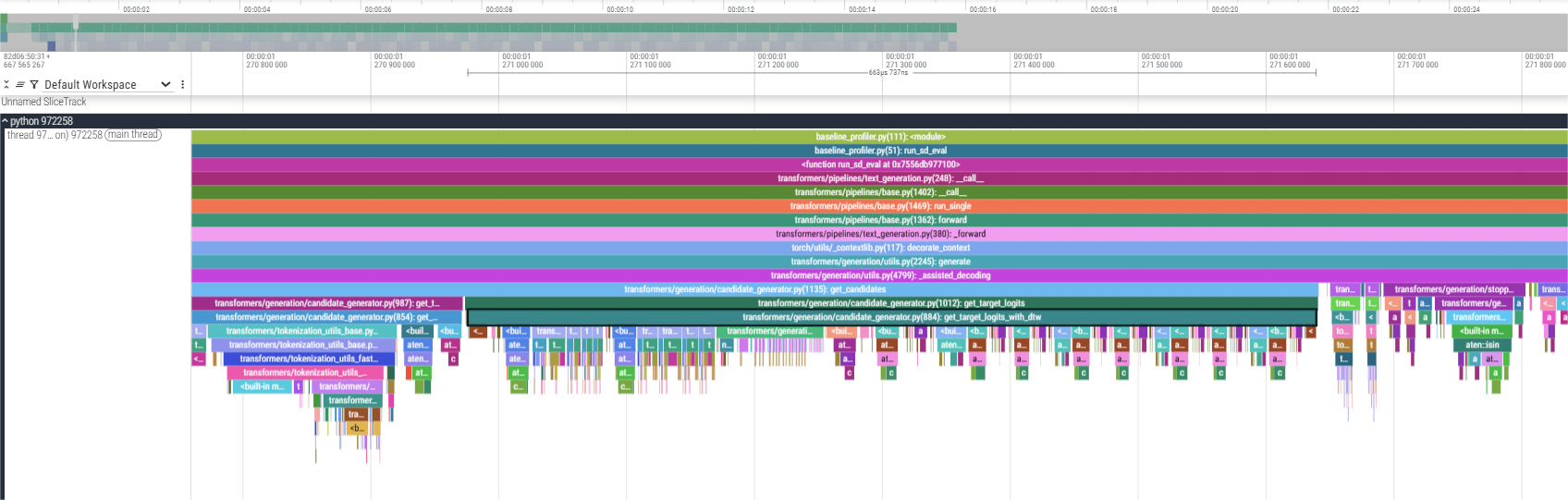}
    \caption{Enlarged view of the process segment for target logits calculation with DTW-related operations in the previous trace.}
    \label{fig:cpu trace amplify}
\end{figure*}

\section{Generation Hyperparameter}
The hyperparameter configurations for text generation of both target models and draft models in our experiment can be seen in Tab.~\ref{tab:cfg}.
\begin{table*}[!ht]
\centering
\small
\begin{tabular}{lcccc}
\toprule
\textbf{Model} & \textbf{Temperature} & \textbf{Top P} & \textbf{Top K} & \textbf{Repetition Penalty} \\
\midrule
Llama-3.1-70B & 0.6 & 0.9 & - & - \\
Deepseek-R1-Distill-Llama-70B & 0.6 & 0.95 & - & - \\
Qwen3-32B & 0.6 & 0.95 & 20 & - \\
Qwen3-30B-A3B & 0.7 & 0.8 & 20 & - \\
Qwen3-0.6B & 0.6 & 0.95 & 20 & - \\
Qwen2.5-0.5B(-Instruct) & 0.7 & 0.8 & 20 & 1.1 \\
\bottomrule
\end{tabular}
\caption{The text generation hyperparameter configurations for target models and draft models in the experiment.}
\label{tab:cfg}
\end{table*}
\label{appendix:trace}

\section{Repetition-N}
The repetition degree indicator is used to indirectly evaluate the diversity of the generated text. It calculates the ratio of the number of n-grams with a frequency higher than 1 in the generated text to the total number of n-grams according to Alg.~\ref{alg:ngram_repetition}. 
\begin{algorithm}[!t]
\caption{N-gram Repetition Calculation}
\label{alg:ngram_repetition}
\begin{algorithmic}[1]
\REQUIRE Text sequence $S$, N-gram order $N = 3$
\ENSURE Repetition rates $R = [r_1, r_2, \dots, r_N]$
\STATE Initialize frequency counters $F_1, F_2, \dots, F_N$
\STATE Initialize repetition rates $R = [0, 0, \dots, 0]$

\FOR{each text $t \in S$}
    \STATE Tokenize $t$ into word sequence $W$
    \FOR{$n = 1$ \TO $N$}
        \STATE Extract all n-grams from $W$ of length $n$
        \STATE Update frequency counts in $F_n$
    \ENDFOR
\ENDFOR

\FOR{$n = 1$ \TO $N$}
    \STATE Count repeated n-grams: $C_{rep} = |\{f \in F_n : f > 1\}|$
    \STATE Count total unique n-grams: $C_{total} = |F_n|$
    \STATE $r_n = C_{rep} / C_{total}$
\ENDFOR

\STATE \RETURN $R \times 100$ \#Convert to percentages
\end{algorithmic}
\end{algorithm}

\section{Token Distance}
\label{appendix:td}
The Edit Distance, often referred to as Levenshtein Distance, is a measure of the similarity between two strings (tokens). It is defined as the minimum number of single-character edits (insertions, deletions, or substitutions) required to change one token into the other.

\subsection{1. Symbol Definition}

Let the source token be denoted by $s = s_1s_2...s_m$, and the target token be denoted by $t = t_1t_2...t_n$.

The edit distance between the first $i$ characters of $s$ and the first $j$ characters of $t$ is denoted by $D(i, j)$. The goal is to compute $D(m, n)$.

The cost of a substitution operation is defined as:
$
\text{cost}_{sub}(s_i, t_j) =
\begin{cases}
    0 & \text{if } s_i = t_j \\
    1 & \text{if } s_i \neq t_j
\end{cases}
$
The cost for a deletion or an insertion is always 1.

\subsection{2. Calculation Process}

We use a dynamic programming approach to compute the distance. A matrix $D$ of size $(m+1) \times (n+1)$ is constructed.
\paragraph{Step 1: Initialization}
The first row and the first column of the matrix are initialized to represent the edits needed to transform a prefix into an empty string, or an empty string into a prefix.
\begin{align*}
    D(i, 0) &= i \quad \text{for } i = 0, \dots, m \\
    D(0, j) &= j \quad \text{for } j = 0, \dots, n
\end{align*}

\paragraph{Step 2: Recurrence Relation}
For every $i$ from 1 to $m$ and every $j$ from 1 to $n$, the value of $D(i, j)$ is calculated as the minimum of three possible operations:
$$
D(i, j) = \min
\begin{cases}
    D(i-1, j) + 1 \quad\text{(deletion)} \\
    D(i, j-1) + 1 \quad\text{(insertion)} \\
    D(i-1, j-1) + \text{cost}_{sub}(s_i, t_j) & \\\text{(substitution)}
\end{cases}
$$

\paragraph{Step 3: Final Result}
The edit distance between the entire token $s$ and token $t$ is the value in the last cell of the matrix:
$
\text{distance}(s, t) = D(m, n)
$

\section{Trace Profiling}

Tokenization, logic operation, etc., these operations performed on the CPU may cause GPU synchronization, which will cause a significant performance degradation. In Fig. \ref{fig:cpu trace} and \ref{fig:cpu trace amplify}, we analyzed the CPU and GPU trace and confirmed that the introduction of DTW did not cause any performance bottlenecks. The blocking time introduced by TokenTiming in one decoding cycle is only 663 $\mu s$, which is insignificant compared to other parts.
\vspace{0.4cm}
\section{Domain-specific Models}

We conducted an additional experiment explicitly targeting specialized vocabulary, with results shown in Tab.~\ref{tab:ds}. We fixed Qwen2.5-0.5B as the draft model and selected two domain-specific target models—-BioMistral-7B for the medical domain and Qwen2.5-Coder-14B for the code domain—-both of which exhibit systematically lower vocabulary overlap with general-purpose drafts. This setup directly evaluates whether TokenTiming remains effective under realistic domain-specific conditions.
\begin{table}[!t]
\centering
\small
\begin{tabular}{c|cc}
\hline
Target & BioMistral-7B & Qwen2.5-Coder-14B \\ 
\hline
Draft & Qwen2.5-0.5B & Qwen2.5-0.5B \\ 
\hline
TLI & 1.15$\times$  & 1.09$\times$ \\ 
\hline
TokenTiming & \textbf{1.31$\times$} & \textbf{1.25$\times$} \\ 
\hline
\end{tabular}
\vspace{0.4cm}
\caption{Speedup on Domain-specific Models}
\label{tab:ds}
\end{table}
\section{Settings of $w$}
Since speculative decoding generates only a bounded number of draft tokens per decoding step, the window size $w$ naturally resides within a finite search space. Our procedure for determining $w$ is as follows. We first perform DTW alignment without any window constraint and collect the offset distribution. As shown in Fig. \ref{fig:cdf}, we then identify, for each model pair, the offset at which the CDF reaches 0.9. These values define a narrowed and more representative search space for $w$. Within this reduced space, we evaluate candidate window sizes and select the setting that yields the highest mean speedup across model pairs. This procedure leads to the global configuration $w=8$ used in our experiments. In practice, $w$ can also be tuned adaptively within the same search space—larger $w$ for model pairs with a later CDF plateau and smaller $w$ for those with an earlier plateau.

\end{document}